\documentclass[11pt]{article}

\usepackage[]{acl}

\usepackage{times}
\usepackage{latexsym}

\usepackage[T1]{fontenc}

\usepackage[utf8]{inputenc}

\usepackage{microtype}

\usepackage{hyperref}
\usepackage{amsmath} 
\usepackage{amssymb} 
\usepackage{graphicx} 
 \usepackage{xcolor}
\usepackage{threeparttable}
\usepackage{pifont}
\usepackage{multirow}
\usepackage{subfigure}
\usepackage{graphicx}
\usepackage{booktabs}
\usepackage{enumitem}
\usepackage{algorithm}
\usepackage{algpseudocode}
\usepackage{amsmath}
\usepackage{graphicx} 
\usepackage{mdframed}
\usepackage{relsize}
\usepackage{ulem}
\usepackage{booktabs}
\usepackage{graphicx}
\usepackage{array}
\usepackage{colortbl} 
\usepackage{array}    

 \usepackage{amsmath}

\usepackage{csquotes}
\usepackage{epigraph}
\usepackage{multirow} 
\usepackage{geometry}
\usepackage{array}
\usepackage{longtable}
\usepackage{float}
\usepackage{mdframed}
\usepackage{xcolor}
\definecolor{deepyellow}{rgb}{0.8, 0.7, 0.0} 
\definecolor{deepgreen}{rgb}{0.0, 0.7, 0.0} 

\definecolor{chatgpt_c}{RGB}{121,147,210}
\definecolor{src_c}{RGB}{238,154,189}
\definecolor{tgt_c}{RGB}{139,212,209}
\definecolor{src_tgt_c}{RGB}{149,149,149}
\definecolor{ibut_c}{RGB}{247,182,95}

\definecolor{chinese_red}{RGB}{230,239,255}
\definecolor{chinese_red_small}{RGB}{252,255,230}
\definecolor{chinese_brown}{RGB}{246,230,255}

\definecolor{enity}{RGB}{255,128,0}
\definecolor{action}{RGB}{102,178,255}

\definecolor{red_word}{RGB}{255,0,0}

 \usepackage{booktabs}

\usepackage{textcomp}
\usepackage{graphicx}

\definecolor{win}{RGB}{165,127,183} 
\definecolor{tie}{RGB}{204,161,189}
\definecolor{loss}{RGB}{229,207,221}

\usepackage{inconsolata}
\NewDocumentCommand{\bai}{ mO{} }{\textcolor{red}
{\textsuperscript{\textit{bai}}\textsf{\textbf{\small[#1]}}}}
\NewDocumentCommand{\andong}{ mO{} }{\textcolor{blue}
{\textsuperscript{\textit{andong}}\textsf{\textbf{\small[#1]}}}}
\title{Make Imagination Clearer! Stable Diffusion-based Visual Imagination for Multimodal Machine Translation}

\author{
Andong Chen\hspace{0.5mm}, 
 Yuchen Song\hspace{0.5mm}, 
\textbf{Kehai Chen}\hspace{0.5mm}, 
 \textbf{Muyun Yang}\hspace{0.5mm},
 \textbf{Tiejun Zhao}\hspace{0.5mm}, 
 \textbf{Min Zhang}\hspace{0.2mm}\hspace{1.5mm} \\
 School of Computer Science and Technology, Harbin Institute of Technology, China\\
  ands691119@gmail.com, 2021113318@stu.hit.edu.cn  \\
  \{chenkehai, baixuefeng, yangmuyun, tjzhao, zhangmin2021\}@hit.edu.cn, 
}

\begin{document}
\maketitle
\begin{abstract}
Visual information has been introduced for enhancing machine translation (MT), and its effectiveness heavily relies on the availability of large amounts of bilingual parallel sentence pairs with manual image annotations.
In this paper, we introduce a stable diffusion-based imagination network into a multimodal large language model (MLLM) to explicitly generate an image for each source sentence, thereby advancing the multimodel MT.
Particularly, we build heuristic human feedback with reinforcement learning to ensure the consistency of the generated image with the source sentence without the supervision of image annotation, which breaks the bottleneck of using visual information in MT.
Furthermore, the proposed method enables imaginative visual information to be integrated into large-scale text-only MT in addition to multimodal MT.
Experimental results show that our model significantly outperforms existing multimodal MT and text-only MT, especially achieving an average improvement of more than 14 BLEU points on Multi30K and MSCOCO multimodal MT benchmarks.

\end{abstract}

\section{Introduction}

Large Language Models (LLMs) have recently demonstrated exceptional comprehension and generation abilities across a wide range of tasks, particularly in translation \citep{tyen2023llms,liang2023encouraging,DBLP:journals/corr/abs-2303-16104,DBLP:journals/corr/abs-2308-14186,zhang-etal-2024-paying,chen2024benchmarking,DBLP:journals/corr/abs-2410-12543,DBLP:journals/corr/abs-2311-07919}. LLM-based machine translation (LLM-MT) methods generally map the source text directly to the target text \cite{hendy2023good,jiao2023chatgpt,le2023bloom,DBLP:conf/wmt/IyerCB23,DBLP:journals/corr/abs-2311-02851,zhao2024review}, while professional human translators often imagine visual information when translating source texts \cite{hubscher2020translation,bang1986imagination,DBLP:conf/naacl/LongWL21,DBLP:conf/ijcnlp/ElliottK17}.
\begin{figure}[!th]\centering
\includegraphics[scale=0.4]{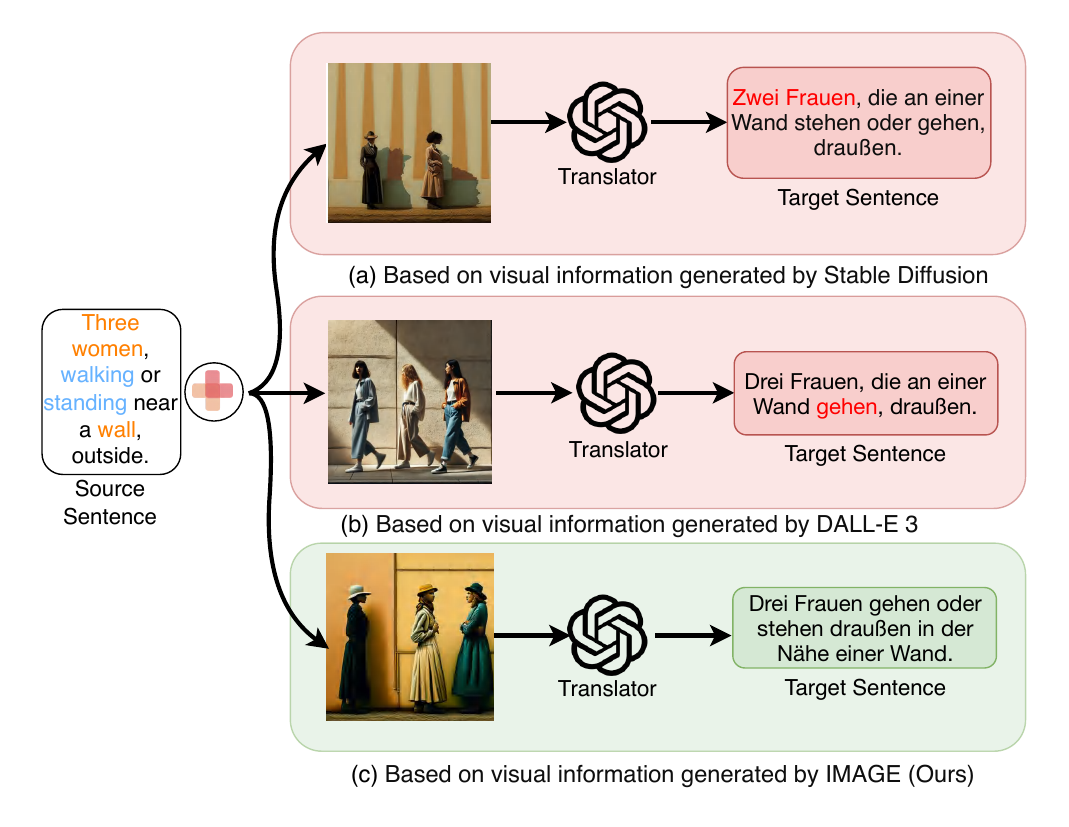} 
\caption{Illustration of the LLMs translation paradigm based on visual information. Figure a: The generated image does not include information about “\textcolor{enity}{three women},” and Figure b: The generated image lacks “\textcolor{action}{standing}” information. These issues led to the translation error.}
\label{intro}
\end{figure}
The process of imagining involves creating scenes, relationships between objects, and commonsense details within the translation text. Therefore, generating such content is crucial for ensuring high-quality translation as it helps capture subtle nuances accurately \cite{DBLP:conf/acl/YaoW20,DBLP:conf/mm/LinMSYYGZL20,DBLP:conf/cvpr/SigurdssonANSMC20,DBLP:journals/tmm/SongCJLXH22}. Although multiple previous works in multimodal machine translation have attempted similar approaches\cite{DBLP:conf/naacl/LongWL21,DBLP:conf/ijcnlp/ElliottK17,DBLP:conf/acl/HitschlerSR16}, they still face limitations such as insufficient model capacity, the requirement for image-text annotated training data, and poor quality of generated images.

To address these issues, we propose a framework called \textbf{IMAGE}, which stands for \textbf{I}magination-Based End-to-End \textbf{M}ultimodal L\textbf{A}rge Langua\textbf{G}e Mod\textbf{E}l Machine Translation Framework. IMAGE first generates corresponding visual information (image) from the source text, and then uses both source text and visual information to produce translation results through LLM. Current mainstream visual information generation methods (such as diffusion models \cite{DBLP:conf/nips/Du0Q023,DBLP:conf/acl/TangLPJYKSLT23,DBLP:conf/aaai/LiuL24,DBLP:conf/cvpr/Liu0CJH24}) often struggle to generate complex scenes based on language descriptions, impacting translation performance, as shown in Figures 1(a) and (b). To ensure that the generated visual information accurately represents the source text, we heuristically build a supervisory signal based on human feedback to enhance the consistency of generated visual content with the source sentence, further improving translation performance, as illustrated in Figure 1(c).

Our framework was evaluated on the standard Multimodal Machine Translation (MMT) dataset Multi30K and the general Neural Machine Translation (NMT) dataset WMT24. Extensive experimental results confirm that the IMAGE framework based on visual imagination outperforms text-only LLM approaches. Additionally, through ablation experiments, we verified the necessity of each component in the IMAGE framework. Furthermore, analysis experiments and case studies reveal a positive correlation between the consistency of visual imagination with the text and translation performance. In summary, our contributions are as follows:

\begin{itemize}
    \item We are the first to propose an end-to-end multimodal machine translation framework leveraging the visual imagination capabilities of LLMs. Our goal is to inspire the translation community to further integrate LLMs and multimodality into future translation research.

    \item Our framework uses human feedback RL during training, eliminating the need for annotated image-text data and reducing annotation costs.
    
    \item Our model demonstrates significant performance improvements on general and multimodal translation benchmarks compared to traditional multimodal translation methods and text-only LLM-MT.
    
\end{itemize}

\begin{figure*}[!th]\centering
\includegraphics[scale=1.70]{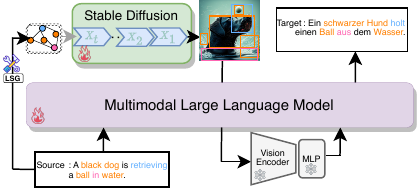} 
\caption{\textbf{Overview of our IMAGE framework.} The process involves first generating visual information of the translation input sentence using a diffusion model. Next, the translation result is obtained via LLM, informed by the generated visual information and translation of the original input sentence.}
\label{framework}
\end{figure*}

\section{Background}

\subsection{Multimodal Large Language Model}
Currently, multimodal large models consist of three main components: a Large Language Model (LLM), an image encoder, and a projector. The \(\mathbf{LLM}\) is responsible for modeling the joint probability distribution \( p_\theta(\mathbf{w}) \) of a sequence \( \mathbf{w} = \left\{\mathbf{w}_t\right\}_{t=1}^T \), where \( T \) is the sequence length and \( \theta \) represents the model parameters. The generation process of each token \( \mathbf{w}_t \) in the LLM is modeled :

\begin{equation}
p_\theta(\mathbf{w}) = \prod_{t=1}^T p_\theta\left(\mathbf{w}_t \mid \mathbf{w}_{<t}\right).    
\end{equation}

For the image encoder, the input sequence contains \( K \) ordered images \( \boldsymbol{I} = \left\{I_k\right\}_{k=1}^K \). Each image \( I_k \) is processed through a vision encoder, such as a CLIP-like encoder \( \mathcal{E}_\phi(\cdot) \), which generates patch embeddings to obtain the image representation signals. These representations are then encoded by the projector \( \mathcal{P}_\zeta \) (such as a linear layer), as described by \citealp{DBLP:conf/nips/AlayracDLMBHLMM22} into visual embeddings \( \boldsymbol{V}_k = \left\{\mathbf{v}_{\ell}\right\}_{\ell=1}^L \) of length \( L \).

Here, \( K(t) \) refers to the image index used before generating the \( t \)-th word token. Maximum likelihood estimation (MLE) aims to minimize the model’s loss function to optimize the parameters \( \theta \), \( \phi \), and \( \zeta \), thereby aligning the generated sequence as closely as possible with the given data. The loss function is written as:

\begin{equation}\small
\mathcal{L}_{\mathrm{MLLM}}(\Theta, \mathbf{w}, \boldsymbol{I}) := 
-\mathbb{E}_t \left[ \log p_{\Theta}\left( \mathbf{w}_t \mid \mathbf{w}_{<t}, \boldsymbol{V}_{<K(t)} \right) \right],
\end{equation}

\begin{equation}
\boldsymbol{V}_{K(t)}=\mathcal{P}_\zeta \circ \mathcal{E}_\phi\left(I_{K(t)}\right).
\end{equation}

\subsection{Scene Graph Representation}
In MMT, the data availability is represented as $<x, z>\in<X, Z>$, where $X$ denotes the source-side sentences and $Z$ represents the paired visual images. Scene Graph represents the semantic relationships between objects in text (LSG) or visual (VSG) information. We define the LSG and VSG as \( LSG = (N_L, E_L) \) and \( VSG = (N_V, E_V) \). The set \( N_L \) and \( N_V \) represent the entity nodes in sentences or visual images, respectively, inculding  the head entity (\( h^l \) and \( h^v \)) and the tail entity (\( t^l \) and \( t^v \)), where \( l \in L \) and \( v \in V \). The sets \( E_L \) and \( E_V \) represent the relations (\( r^l \) and \( r^v \)) connecting these nodes in \( N_L \) and \( N_V \).

\subsection{Diffusion Models}
Diffusion models (DMs) are probabilistic generative models that learn the latent structure of data $\mathbf{x}=\left\{\mathbf{x}_t\right\}_{t=1}^T$ through continuous-$T$-timestamps information diffusion. DMs gradually add Gaussian noise to an image $x_0$ until attaining $\mathbf{x}_T \sim \mathcal{N}(\mathbf{0}, \mathbf{I})$. This noise injection process (the forward process) is formalized as Markov chain $q\left(\mathbf{x}_{1: T} \mid \mathbf{x}_0,c\right)=$ $\prod_{t=1}^T q\left(\mathbf{x}_t \mid \mathbf{x}_{t-1},c\right)$, where $c$ where \( x_0 \) is the sample dataset and \( c \) is the corresponding context. The forward process is written as. 
\begin{equation}
q\left(\mathbf{x}_t \mid \mathbf{x}_0\right)=\mathcal{N}\left(\mathbf{x}_t ; \sqrt{\bar{\alpha}_t} \mathbf{x}_0,\left(1-\bar{\alpha}_t\right) \mathbf{I}\right),
\end{equation}
where $\bar{\alpha}_t=\prod_{i=1}^t \alpha_i$. And $\mathbf{x}_t=\sqrt{\bar{\alpha}} \mathbf{x}_0+\sqrt{1-\bar{\alpha}_t} \epsilon_t$, where $\epsilon_t \sim \mathcal{N}(\mathbf{0}, \mathbf{I})$.

Reversing the forward process can be accomplished by
training a neural network $\mu_\theta (x_t, c, t)$ with the following objective:

\begin{equation}
\label{ddpm}
\resizebox{\linewidth}{!}{$
    L_{DDPM}(\theta) = \mathbb{E}_{(x_0,c) \sim p, t \sim U \{0,T\}, x_t \sim q} \left[ \| \tilde{\mu}(x_0, t) - \mu_\theta (x_t, c, t) \|^2 \right],
    $}
\end{equation}
where $\tilde{\mu}$ is the posterior mean of the forward process, a weighted average of $x_{0}$ and $x_{t}$. This objective is justified as maximizing a variational lower bound on the log-likelihood of the data \cite{DBLP:conf/nips/HoJA20}.

\section{Proposed Framework: IMAGE}

\subsection{Framework Overview}

Our proposed framework, IMAGE, incorporates visual signals to enhance the performance of large models in multilingual translation tasks. Additionally, to ensure that the entity relationships within the generated visual information remain consistent with input sentences, we adopt an alignment human feedback learning approach.

Figure 2 provides an overview of IMAGE. The following subsections detail three key components: the end-to-end multimodal machine translation framework (Section \ref{mt_methods}), alignment human feedback learning (Section \ref{aligment_method}), and the model training process (Section \ref{train_method}).

\subsection{End-to-End Multimodal Machine Translation Framework}
\label{mt_methods}
IMAGE is built upon a causal decoder architecture LLM \(p_{\theta}\), such as Vicuna \cite{vicuna2023}. IMAGE adopts OpenAI’s CLIP-Large  \cite{DBLP:conf/icml/RadfordKHRGASAM21} as the visual encoder \( \mathcal{E}_\phi(\cdot) \), followed by a linear layer \( \mathcal{P}_\zeta \) for visual embedding projection \cite{DBLP:conf/iclr/DongHPQGYZSZWK024}. To generate images, we utilize Stable Diffusion (SD) \cite{DBLP:conf/cvpr/RombachBLEO22} as the image decoder, with the condition projector also implemented as a linear layer. Figure \ref{framework} provides an overview of this architecture.

\subsection{Alignment Human Feedback Learning}
\label{aligment_method}

The Alignment Feedback Learning aims to enhance the quality of images generated by diffusion model through alignment between linguistic and visual information. This method comprises two core parts: \textbf{Design Reward Function} and \textbf{Alignment Optimization For Diffusion Model}.

\subsubsection{Design Reward Function} 
\label{rlhf_section}
\begin{figure}[!th]\centering
\includegraphics[scale=0.4900]{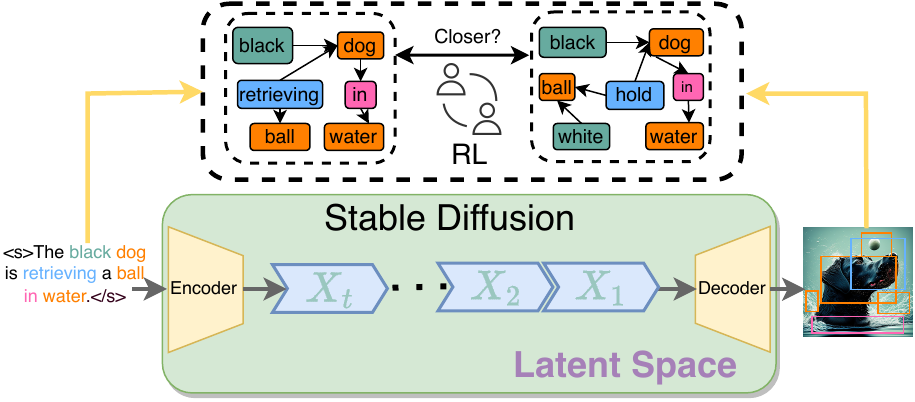} 
\caption{\textbf{RL Training Detail. }The overview of IMAGE, which leverages an alignment feedback learning framework to comprehensively enhance the visual signals performance.}
\label{graph_method_rl}
\end{figure}

To ensure consistency between the translated source sentence and the generated image, the entities and relations in the image need to match those in the source sentence as closely as possible. Based on this, we design a reward function to assess the consistency of the generated image (VSG) to the source sentence (LSG). As shown in Figure \ref{graph_method_rl}, the closer LSG is to VSG, the higher the consistency between the translated source sentence and the generated image. We constructed the reward function to evaluate the consistency of LSG and VSG, with a reward scoring range from 0 to 1. Since human judgments of the consistency between images and descriptive texts are also based on the analysis of entities and their relationships,  this task constitutes reinforcement learning from human feedback \cite{ouyang2022training,DBLP:conf/nips/ChristianoLBMLA17,DBLP:journals/corr/abs-1909-08593}.

\vspace{-0.3cm}

For LSG and VSG generation, we utilize two off-the-shelf SG parsers to obtain LSG and VSG separately (as detailed in \textsection \ref{off-the-shelf-tool}). Due to the differing number of triples in LSG and VSG, we designed a structured similarity calculation method to measure their consistency. For each triple in LSG, we calculate its similarity with each triple in VSG and select the highest score as the matching degree for that triple: 
\vspace{-0.3cm}
\begin{equation}
\resizebox{\linewidth}{!}{$
Score\left(\mathrm{LSG}_i, VSG\right)=\max \left(Sim\left(\mathrm{LSG}_i, \mathrm{VSG}_1\right), \cdots, Sim\left(\mathrm{LSG}_i, \mathrm{VSG}_n\right)\right),
$}
\end{equation}
\vspace{-0.5cm}
\begin{equation}
\resizebox{\linewidth}{!}{$
\text{Sim}(LSG_1, VSG_1) = \frac{SIM(h^{l}_1, h^{v}_1) + SIM(r^{l}_1, r^{v}_1) + SIM(t^{l}_1, t^{v}_1)}{3},
$}
\end{equation}
where  $n$ represents number of VSG sets, \( h^{l} \) and \( h^{v} \) are the head entities, \( t^{l} \) and \( t^{v} \) are the tail entities, \( r^{l} \) and \( r^{v} \) are the relations, and $SIM$ is off-the-shelf similarity of text model (as detailed in \textsection \ref{off-the-shelf-tool}). Finally, the consistency reward score between sentences and images is the average score of all text triples:
\begin{equation}\small
    r(x_0,c) = \frac{1}{N} \sum_{i=1}^{N} Score(LSG_i,VSG),
    \label{reward_score_equaction}
\end{equation}
where \(c\) and $x_0$ denote the LSG of the source sentence and generated image, respectively.

\subsubsection{Alignment Optimization For Diffusion Model}
We assume a pre-existing diffusion model, which may be pretrained. Given a fixed sampler, the diffusion model induces a sample distribution \( p_\theta(x_0 | c) \). The objective of denoising diffusion reinforcement learning (RL) is to maximize a reward signal \( r \) defined on the samples and contexts:
\begin{equation}
\mathcal{L}_{\text{IMAGERL}}(\theta) = \mathbb{E}_{c \sim p(c), x_0 \sim p_\theta(x_0 | c)} [r(x_0, c)], 
\label{loss_image}
\end{equation}
for a context distribution \( p(c) \) of our choosing.

To improve the alignment between generated images and text, we need to optimize $L_{\text{IMAGERL}}$. In general, we can use the denoising loss $L_{\text{DDPM}}$ (Equation \ref{ddpm}), but with training data $x_0 \sim p_\theta (x_0 | c)$ and an added weighting that depends on the reward $r(x_0, c)$. We refer to this general class of algorithms as Denoising Diffusion Policy Optimization (DDPO) \cite{DBLP:conf/iclr/BlackJDKL24}, framing the training of the diffusion model as a Markov Decision Process (MDP) and performing multi-step optimization for fine-tuning.

\begin{table*}[!ht]\centering
\scalebox{0.70}{ 
\begin{tabular}{@{}lccccccc@{}}
\toprule
\textbf{Language} & \multicolumn{3}{c}{\textbf{English $\rightarrow$ German}} & \multicolumn{3}{c}{\textbf{English $\rightarrow$ French}} & \textbf{Average} \\ \midrule
\textbf{Testset}  & Test2016        & Test2017       & MSCOCO         & Test2016       & Test2017       & MSCOCO         & \\ \midrule
\textbf{Metric}    & \multicolumn{6}{c}{BLEU $\uparrow$ / COMET $\uparrow$ / BLEURT $\uparrow$} &  \\ \midrule
\rowcolor{gray!30}  
\multicolumn{1}{c}{}              & \multicolumn{6}{c}{\textbf{Traditional MMT}} & \\ \midrule
Soul-Mix        & 44.2/------/------        & 37.1/------/------       & 34.2/------/------       & 64.7/------/------       & 57.4/------/------       & 49.2/------/------       & 47.8/------/------      \\
RG-MMT-EDC      & 42.2/------/------        & 33.4/------/------       & 30.0/------/------       & 62.9/------/------       & 55.8/------/------       & 45.1/------/------       & 44.9/------/------       \\
WRA-guided      & 39.3/------/------        & 32.3/------/------       & 28.5/------/------       & 61.8/------/------       & 54.1/------/------       & 43.4/------/------       & 43.2/------/------       \\
ImagiT          & 38.6/------/------        & 32.1/------/------       & 29.7/------/------       & 60.8/------/------       & 52.8/------/------       & 42.5/------/------       & 42.7/------/------       \\ 
Imagination     & 39.7/------/------        & 32.3/------/------       & 28.5/------/------       & 61.8/------/------       & 54.1/------/------       & 43.4/------/------       & 43.3/------/------       \\
\midrule
\rowcolor{gray!30} 
\multicolumn{1}{c}{}              & \multicolumn{6}{c}{\textbf{Open-source LLMs based on Text}} & \\ \midrule
Llama3-8B       & 30.1/69.5/56.6  & 24.2/66.4/53.0 & 21.9/62.6/47.8 & 50.2/77.8/61.1 & 40.4/72.8/53.3 & 34.5/70.7/49.9 & 33.6/69.9/53.6 \\
Alpaca-7B       & 38.5/77.2/66.2  & 34.3/76.5/65.9 & 30.9/72.4/61.5 & 59.2/82.5/70.2 & 51.4/79.4/68.3 & 42.6/77.2/62.9 & 42.8/77.5/65.8 \\
Vicuna-7B       & 32.9/75.9/63.5  & 28.0/75.4/63.5 & 26.1/70.3/57.7 & 46.5/81.4/64.8 & 43.8/82.4/66.3 & 39.3/78.6/61.0 & 36.1/77.3/62.8 \\
Tower-7B\textbf{*}        & 22.1/52.1/34.2  & 13.7/45.5/25.8 & 16.3/48.6/31.5 & 24.5/55.9/31.7 & 20.8/50.1/25.7 & 22.5/52.1/29.1 & 20.0/50.7/29.7 \\
ALMA-7B\textbf{*}         & 23.1/66.4/59.1   & 18.9/66.3/57.8 & 13.7/62.1/55.6 & 21.4/67.0/52.6 & 17.4/65.5/50.8 & 17.9/65.3/52.8 & 18.7/65.4/54.8 \\
ALMA-R-13B\textbf{*}      & 29.1/71.8/59.4   & 24.8/71.8/60.5 & 23.9/68.2/57.8 & 27.4/73.7/52.7 & 24.4/74.5/54.6 & 29.2/72.8/54.9 & 26.5/72.1/56.7 \\ \midrule
\rowcolor{gray!30} 
\multicolumn{1}{c}{}              & \multicolumn{6}{c}{\textbf{Open-source LLMs based on Text \& Image}} & \\ \midrule
DreamLLM        & 27.2/74.8/67.4  & 19.5/73.5/65.9 & 19.3/69.4/62.5 & 36.9/81.1/68.3 & 34.7/80.6/67.9 & 36.6/79.2/66.5 & 29.0/76.4/66.4 \\
\textbf{IMAGE}   & \textbf{45.3/83.1/78.1}  & \textbf{38.6/81.9/76.8} & \textbf{37.5/78.8/74.6} & \textbf{67.5/88.3/81.2} & \textbf{61.5/86.6/78.8} & \textbf{49.3/82.5/72.6} & \textbf{49.9/83.5/77.0} \\ \bottomrule
\end{tabular}}
\caption{Main translation results from the Multi30K benchmark, with BLEU, COMET, and BLEURT scores. The bolded results indicate the highest statistically significant scores (p-value $< 0.01$ in the paired t-test against all compared methods). \textbf{*} indicates that no fine-tuning was performed on the Multi30K test set.}
\label{multi30k_main}
\end{table*}

\subsection{Model Training}
\label{train_method}

\textbf{Training of Diffusion Models with RL:} The training objective is to maximize cumulative rewards, improving the alignment between images and text in Equation \ref{loss_image}. We use policy gradient estimation to optimize the model parameters. With access to likelihoods and likelihood gradients, we can make direct
Monte Carlo estimates of $\nabla_\theta \mathcal{L}_{IMAGERL}$. The process uses the score function policy gradient estimator,
also known as the likelihood ratio method or REINFORCE \cite{DBLP:journals/ml/Williams92,DBLP:journals/jmlr/MohamedRFM20}:  
\begin{equation}\small
    \nabla_\theta \mathcal{L}_{IMAGERL} = \mathbb{E}\left[\sum_{t=0}^{T} \nabla_\theta \log p_\theta(x_{t-1} | x_t, c) \, r(x_0, c)\right].
\end{equation}

\textbf{Ordered Learning Implementation:} In the initial stage, each of the above learning objectives will be executed separately in a certain order to maintain a stable and effective IMAGE system. We first perform \(\mathcal{L}_{\mathrm{IMAGERL}}\). After training Diffusion Models, we train LLM with the loss  \(\mathcal{L}\) which is the combination of \(\mathcal{L}_{\mathrm{IMAGERL}}\) and \(\mathcal{L}_{\mathrm{MLLM}}\): 
\begin{equation}
    \mathcal{L} = \frac{\mathcal{L_{\mathrm{MLLM}}}}{\mathcal{L_{\mathrm{MLLM}}}^{constant}}+\frac{\mathcal{L_{\mathrm{IMAGERL}}}}{\mathcal{L_{\mathrm{IMAGERL}}}^{constant}},
\end{equation}
where \(constant\) refers to the loss value treated as a constant.

\section{Experiment Setup}

\subsection{Data and Training Setting}
\textbf{Dataset}: We conduct experiments on two MT benchmarks: Multi30K\cite{DBLP:conf/acl/ElliottFSS16} and WMT24 test set \cite{kocmi-etal-2024-findings}. Dataset details are in Appendix \ref{data_appendix}.

\textbf{Training Setting}: Details of our training setting and off-the-shelf tools are in Appendix \ref{off-the-shelf-tool}.

\subsection{Comparing Systems}
We used two types of baseline methods:

(\romannumeral1) \textbf{Traditional Multimodal Machine Translation models (MMT)}, including Soul-Mix\cite{cheng-etal-2024-soul}, RG-MMT-EDC\cite{DBLP:journals/talip/TayirL24}, WRA-guided\cite{DBLP:journals/taslp/ZhaoKKC22}, Imagination\cite{DBLP:conf/ijcnlp/ElliottK17} and ImagiT\cite{DBLP:conf/naacl/LongWL21}. These MMT baselines take the source language sentence as textual input while utilizing the image as visual input. They have completed training on the Multi30k training dataset and reached convergence. The results are cited from the reported data in the paper.


(\romannumeral2) \textbf{Open-source Large language models}, including Llama3-8B, Alpaca-7B, Vicuna-7B, Tower-7B, ALMA-7B, ALMA-R-13B, and DreamLLM. Among them, Llama3-8B\cite{llama3modelcard}, Alpaca-7B\cite{Bommasani2021FoundationModels}, and Vicuna-7B\cite{vicuna2023} are models widely used for multilingual tasks, all of which exhibit strong instruction-following capabilities. For Tower-7B\cite{tower_llm_2024}, ALMA-7B\cite{xu2023paradigm}, and ALMA-R-13B\cite{xu2024contrastive}, these models were pre-trained and fine-tuned on translation datasets, outperforming ChatGPT in multiple language directions. DreamLLM\cite{DBLP:conf/iclr/DongHPQGYZSZWK024}  is a framework that unifies text and image generation in multimodal Large Language Models.

\subsection{Automatic Evaluation}
In evaluating our translation methodology, we initially employ COMET\footnote{https://huggingface.co/Unbabel/wmt22-comet-da} \cite{rei2022comet} and BLEURT\footnote{https://github.com/lucadiliello/bleurt-pytorch} \cite{sellam2020bleurt} as automatic metrics, aligning with the established standards in LLM-based translation literature \cite{chen-etal-2024-dual,He2023ExploringHT,huang2024aligning}. For traditional translation evaluation, we use BLEU \footnote{https://github.com/mjpost/sacrebleu} \cite{papineni2002bleu}.

\section{Experimental Results}

\subsection{Main Experiment Results on MMT task}

In Table \ref{multi30k_main}, we present the overall experimental results on the classic Multi30K dataset in the MMT field. First, we compare different methods fine-tuned on the same training set. Our method demonstrates significant improvement in translation performance by generating visual information, clearly outperforming text-only translation models based on the same foundational LLM in this task by average 13.7=(12.4+10.6+11.4+21.1+16.7+10)/6 BLEU score, highlighting the critical role of visual information in text translation (consistent with the conclusion in Section \ref{loss_ablation}). Next, we compare our method with traditional multimodal machine translation (MMT) research. Traditional MMT methods, developed over years of study, can make more comprehensive use of annotated image information. However, IMAGE still surpasses these methods, showcasing the potential of multimodal large language models in MT.
\vspace{-0.3cm}

\subsection{Main Experiment Results on General MT}

\textbf{The effectiveness of IMAGE in general domain translation tasks}. In the WMT24 general domain tasks, as shown in Table \ref{wmt24}, IMAGE outperforms other methods across 4 language pairs and 3 evaluation metrics. Specifically, in the general domain, the IMAGE method outperforms Vicuna directly by +3.9 BLEU and +8.2 COMET. This indicates that the visual information enhances the translation ability of LLMs in the general MT task.

\begin{table}[ht]\centering
\scalebox{0.570}{ 
\begin{tabular}{@{}lcccc@{}}
\toprule
           & \textbf{En$\rightarrow$Zh}     & \textbf{En $\rightarrow$ De}     & \textbf{En$\rightarrow$Hi}     & \textbf{En$\rightarrow$Cs}     \\ \midrule
 \rowcolor{gray!30}            & \multicolumn{4}{c}{BLEU $\uparrow$ /COMET $\uparrow$ /BLEURT $\uparrow$ }                \\ \midrule
Llama3-8B  & 11.6/56.8/33.4 &	12.7/54.3/36.9	& 1.2/39.4/31.5	& 3.2/47.9/25.0 \\
Alpaca-7B  & 15.0/54.6/45.7 & 17.1/60.4/56.5 & 2.9/36.7/36.5 & 3.4/53.6/36.7 \\
Vicuna-7B  & 21.8/63.9/36.4 & 23.3/68.2/52.1 & 5.6/49.4/45.0 & 6.7/57.9/45.2 \\
Tower-7B\textbf{*}   & 13.5/55.5/42.8 & 17.2/55.7/47.2 & 2.0/32.1/20.2 & 1.4/42.9/28.9 \\
ALMA-7B\textbf{*}    & 14.8/52.9/33.4 & 17.4/58.1/40.2 & 1.0/31.9/26.9 & 1.7/49.7/32.0 \\
ALMA-R-13B\textbf{*} & 15.2/57.4/37.2 & 18.3/57.2/46.8 & 1.3/34.1/30.9 & 3.5/53.2/45.5 \\
IMAGE       & 26.8/77.6/57.4 & 23.8/73.3/60.8 & 6.2/51.4/47.3 & 16.2/69.9/53.9 \\ \bottomrule
\end{tabular}}
\caption{Main translation results from the WMT24 test set, with BLEU and COMET scores. The bolded results indicate the highest statistically significant scores (p-value $< 0.01$ in the paired t-test against all compared methods). \textbf{*} indicates that no fine-tuning was performed on the WMT24 test set.}
\label{wmt24}
\end{table}

\textbf{The effectiveness of IMAGE in low-resource tasks.} We selected 2 low-resource tasks (En$\rightarrow$Cs, En$\rightarrow$Hi) from WMT24. As observed in Table \ref{wmt24}, current low-resource tasks still pose challenges to LLMs. However, compared to baseline methods, 
IMAGE achieved an average improvement of +14.13 COMET and +3.87 BLEU for En$\rightarrow$Hi, and +19.03 COMET and +12.88 BLEU for En$\rightarrow$Cs, respectively. This suggests that visual information can provide supplementary data for low-resource tasks, thereby enhancing translation performance in low-resource scenarios.

\subsection{Experiment on the Correlation between Reward Scores and MT Performance}

We further investigated the impact of the proposed RL training method on model translation performance. Inspired by \citealp{DBLP:conf/acl/WuKBLK20} and \citealp{DBLP:conf/acl/ZhuSCHWW23}, we conducted a visual analysis on Multi30K (En$\rightarrow$De), using BLEU and Reward scores (calculated as shown in Equation \ref{reward_score_equaction}) as reference metrics. 

\begin{figure}[!th]
\centering
\includegraphics[scale=0.23]{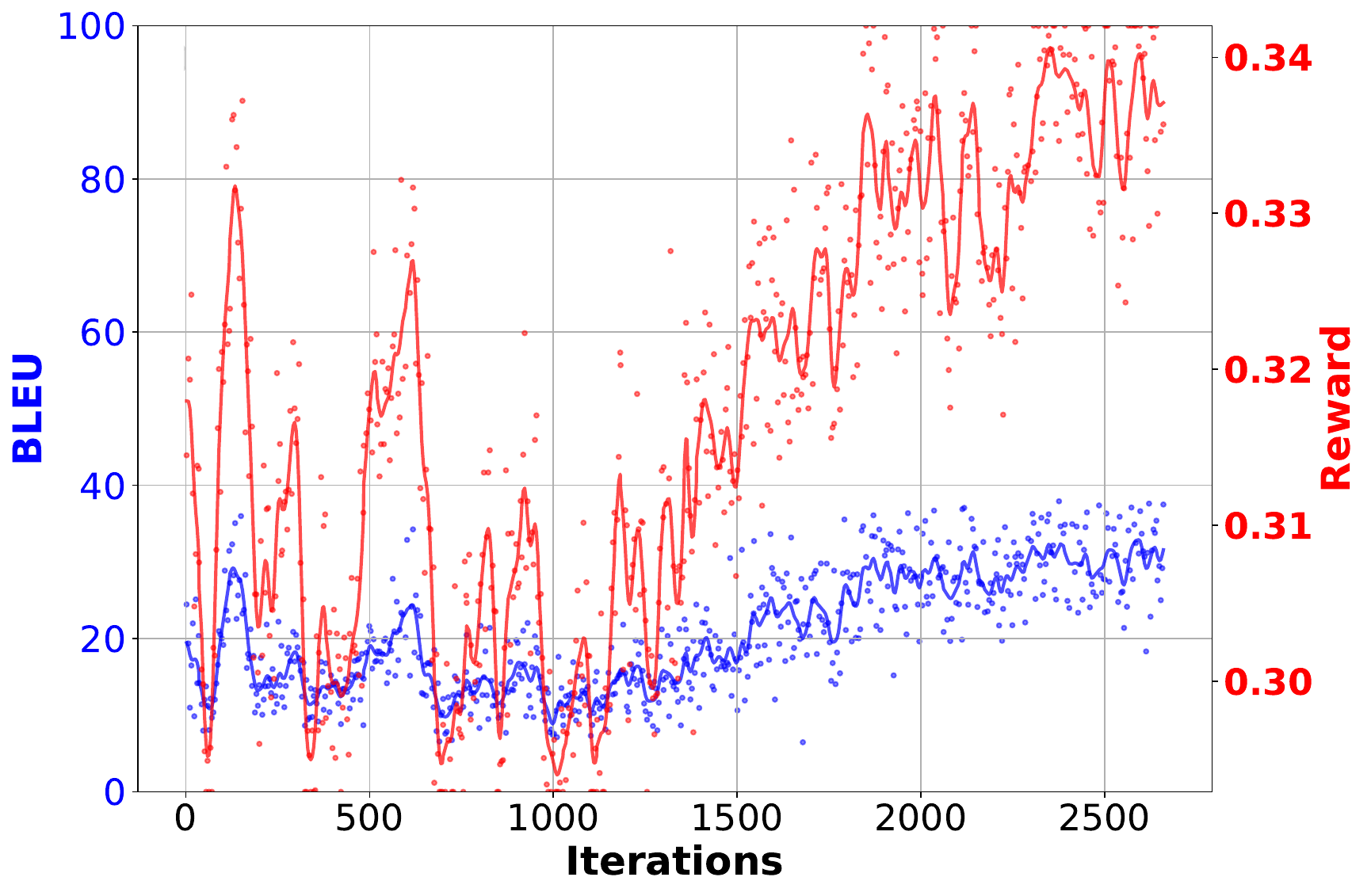} 
\caption{Analysis of the experimental setup for assessing the impact of the Iterative Refinement part on translation performance.}
\label{iter_rw_bleu}
\end{figure}
\vspace{-0.3cm}

Figure \ref{iter_rw_bleu} presents the results of the training phase, where the horizontal axis shows the number of training iterations, and the vertical axes show translation performance (left) and RL Reward scores (right). The results indicate that as training progresses, our method continues to optimize, with Reward scores gradually increasing and translation quality improving. Additionally, the Reward score measures the similarity between LSG and VSG. The experimental results show that as this similarity increases, the generated images align more closely with the source sentences, effectively enhancing translation performance.


\subsection{Ablation Experiment on Loss}
\label{loss_ablation}

In Table \ref{ablation_loss}, we quantify the contribution of each learning strategy through the ablation study. Each learning strategy has a significant impact on overall performance. The training objective aligning visual and source sentence information demonstrates a notable impact, with an average increase of 1.5 scores. Additionally, multilingual text translation showed a more significant effect, with an average increase of 7 BLEU scores. When using these two training objectives together, we observed the most significant performance improvement, with an average increase of 18.4 BLEU scores. These results confirm the long-standing findings in MMT research on the positive influence of visual information on multilingual translation tasks \cite{DBLP:conf/eamt/ZhaoKKC20,DBLP:conf/acl/FangF22,DBLP:conf/acl/ElliottFSS16}.

\begin{table}[!ht]
\centering
\scalebox{0.60}{
\begin{tabular}{@{}llccc@{}}
\toprule
\multicolumn{2}{c}{\textbf{Configuration}} & \multicolumn{3}{c}{\textbf{English $\rightarrow$ German}} \\ \cmidrule(lr){1-2} \cmidrule(lr){3-5}
\textbf{$L_{MLLM}$} & \textbf{$L_{IMAGERL}$} & \textbf{Test2016} & \textbf{Test2017} & \textbf{MSCOCO} \\ \midrule
\ding{55} & \ding{55} & 27.2 / 74.8 / 67.4 & 19.5 / 73.5 / 65.9 & 19.3 / 69.4 / 62.5 \\
\ding{55} & \checkmark & 27.4 / 74.9 / 67.5 & 22.2 / 74.3 / 66.6 & 21.0 / 71.5 / 63.2 \\
\checkmark & \ding{55} & 32.9 / 75.9 / 63.5 & 28.0 / 75.4 / 63.5 & 26.1 / 70.3 / 57.7 \\
\checkmark & \checkmark & 45.3 / 83.1 / 78.1 & 38.6 / 81.9 / 76.8 & 37.5 / 78.8 / 74.6 \\ \bottomrule
\end{tabular}}
\caption{Comparison of configurations with different loss functions ($L_{MLLM}$ and $L_{IMAGERL}$). Metrics are BLEU/COMET/BLEURT.}
\label{ablation_loss}
\end{table}
\vspace{-0.3cm}

\subsection{Ablation Experiment on Module}

In Table \ref{ablation_module}, We conducted ablation studies on Multi30K to assess the role of each component in the IMAGE. Removing the Stable Diffusion model (w/o SD) led to an average BLEU score decrease of 1.7, showing that generated visual information improves multilingual translation. Replacing SD-generated images with real images (w/ RI) caused a 1.8-point drop, indicating SD-generated images provide greater benefits (we will further discuss this phenomenon in Section \ref{section_image_quality}). Removing vision encoder (CLIP features) (w/o VS) resulted in a significant BLEU score decline (45.43/38.6/37.5 without CLIP, compared to 39.2/35.1/33.2 with CLIP), highlighting the importance of vision encoder in aligning vision and text.


\begin{table}[!ht]
\centering
\resizebox{0.47\textwidth}{!}{
\begin{tabular}{@{}lccc@{}}
\toprule
\textbf{Language} & \multicolumn{3}{c}{\textbf{English$\rightarrow$German}} \\ \midrule
\textbf{Testset}  & \textbf{Test2016} & \textbf{Test2017} & \textbf{MSCOCO} \\ \midrule
\textbf{Metrics}  & \multicolumn{3}{c}{BLEU $\uparrow$ / COMET $\uparrow$ / BLEURT $\uparrow$} \\ \midrule
IMAGE             & 45.3/83.1/78.1 & 38.6/81.9/76.8 & 37.5/78.8/74.6 \\
\quad - w/o SD    & 42.9/82.5/77.2 & 37.7/81.4/76.2 & 35.6/78.6/73.9 \\
\quad - w/ RI     & 42.6/82.3/77.0 & 37.9/81.3/76.1 & 35.5/78.7/74.1 \\
\quad - w/o VS    & 39.2/77.7/67.2 & 35.1/77.4/67.0 & 33.2/72.7/61.9 \\ \bottomrule
\end{tabular}}
\caption{Comparison of configurations with different modules. SD, RI, and VS represent Stable Diffusion, Real Image, and Vision Encoder, respectively. Metrics are BLEU, COMET, and BLEURT.}
\label{ablation_module}
\end{table}
\vspace{-0.5cm}

\subsection{Evaluation of Generated Image Quality}
\label{section_image_quality}

To investigate the correspondence between the images generated by IMAGE and the source language sentences, we used the pretrained Stable Diffusion model and IMAGE to generate images, and then calculated the $CLIPScore$ \cite{DBLP:conf/emnlp/HesselHFBC21}. $CLIPScore$ measures the similarity between the image and the source language sentence using the formula: $CLIPScore(c, v) = max(cos(c, v), 0)$, where $c$ and $v$ are the feature vectors from the text encoder and the image encoder of CLIP \cite{DBLP:conf/icml/RadfordKHRGASAM21}, respectively.

The evaluation results in Table \ref{clip_image_quality} show that IMAGE outperforms the pretrained Stable Diffusion model across all datasets. Additionally, IMAGE-generated images exhibit higher similarity to the source language sentences than the original related images in Test2016 and Ambiguous COCO. This confirms that our method generates images that better reflect the source language, enhancing translation tasks.

\begin{table}[!ht]\centering 
\scalebox{0.8}{ 
\begin{tabular}{@{}lccc@{}}
\toprule
\textbf{Language} & \multicolumn{3}{c}{\textbf{English $\rightarrow$ German}} \\ \midrule
\textbf{Testset}  & \textbf{Test2016}   & \textbf{Test2017}   & \textbf{MSCOCO}   \\ \midrule
\textbf{Metrics}  & \multicolumn{3}{c}{CLIPScore $\uparrow$ }      \\ \midrule
Stable Diffusion \includegraphics[width=0.0225\textwidth]{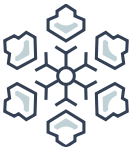}  & 0.72       & 0.72       & 0.71     \\
IMAGE (SD) \includegraphics[width=0.0175\textwidth]{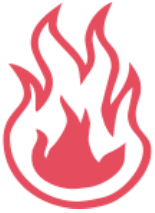}         & \textbf{0.76}       & 0.76       & \textbf{0.75}     \\
Multi30K          & 0.75       & \textbf{0.78}       & 0.74     \\ \bottomrule
\end{tabular}}
\caption{CLIPScore: Similarity between Source Language Sentences and Related Images. \includegraphics[width=0.0225\textwidth]{ice.pdf} indicates Stable Diffusion without fine-tuning. \includegraphics[width=0.0175\textwidth]{fire.pdf} indicates Stable Diffusion fine-tuned with RL (\textsection \ref{rlhf_section}).}
\label{clip_image_quality}
\end{table}
\vspace{-0.3cm}

\begin{figure*}[!th]\centering
\includegraphics[scale=0.35]{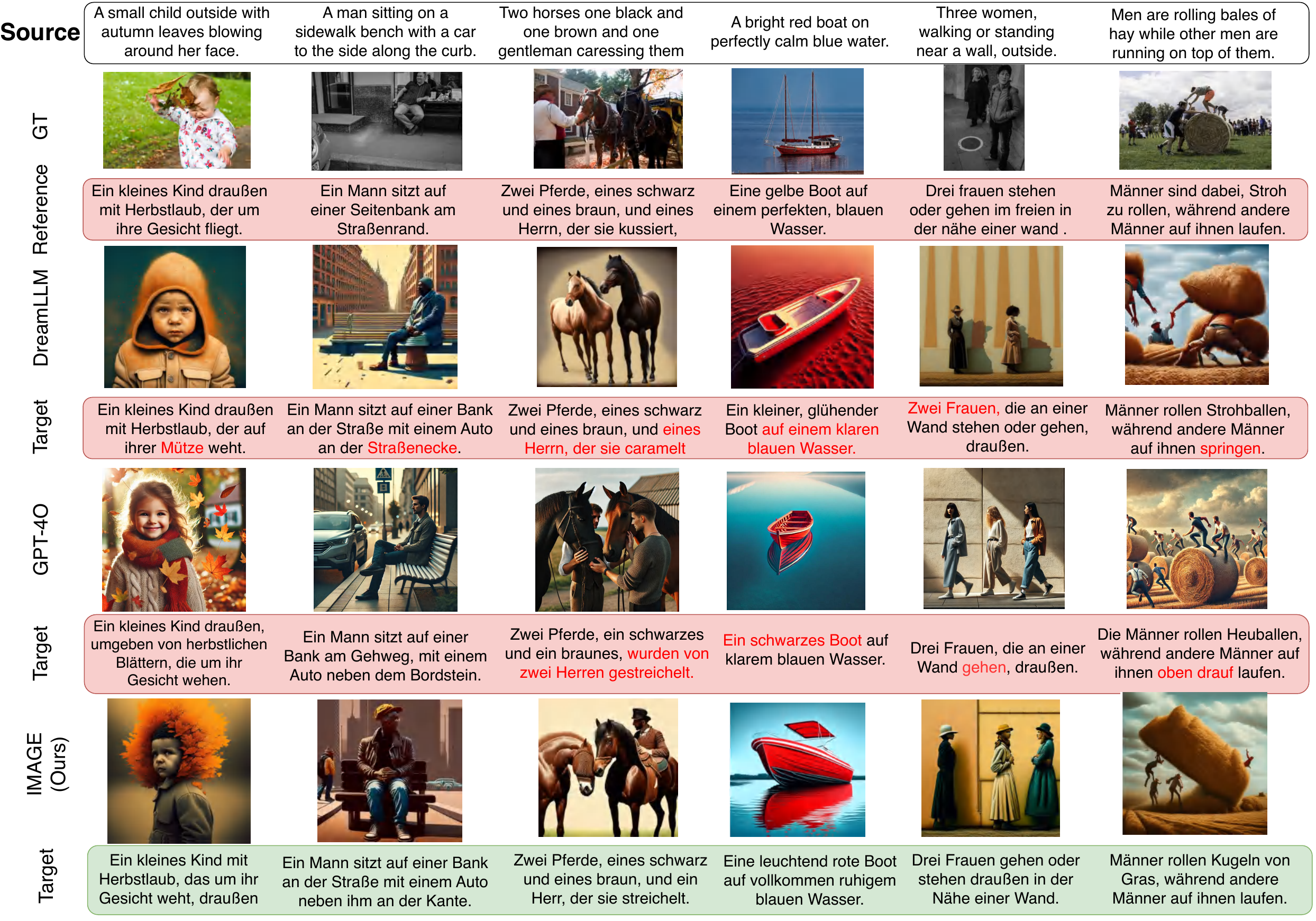} 
\caption{\textbf{Some qualitative results on the comparison of IMAGE against related work on the Multi30K En-De test set.} IMAGE, in addition to high quality image generation, correctly generates the number of given instances in the image and represents the scene more accurately overall. GPT-4O refers to using DALL-E for image generation, followed by GPT-4O model performing translation based on the source sentence and the generated image. \textcolor{red_word}{Red words} indicate the parts with translation errors.}
\label{case_study}
\end{figure*}

We also present some qualitative case study results on the Multi30K En$\rightarrow$De test datas in Figure \ref{case_study}. The results indicate that, compared to Stable Diffusion and OpenAI’s DALL-E 3\footnote{https://openai.com/index/dall-e-3/}, our proposed model generates more accurate images based on the source sentences, leading to higher-quality translation outcomes. A key advantage of the IMAGE model is its ability to generate visuals that correctly represent the number and relationships of object instances as defined by the source sentence, ensuring translation accuracy.

\section{Related Works}
\textbf{MMT Model Architecture}: Multimodal Machine Translation (MMT) aims to enhance machine translation tasks through the aid of visual information \cite{zhang2019bridging}. Since the release of the Multi30K dataset \cite{DBLP:conf/acl/ElliottFSS16}, early research has primarily focused on model architecture design \cite{DBLP:conf/emnlp/ZhouCLY18,DBLP:conf/emnlp/CalixtoL17,DBLP:conf/wmt/HelclLV18}. Subsequent studies, such as those by \citealp{DBLP:conf/acl/YaoW20} and  \citealp{DBLP:conf/acl/YinMSZYZL20}, proposed multimodal encoders that integrate text and visual information during the encoding stage. \citealp{DBLP:conf/acl/IveMS19} and \citealp{DBLP:conf/mm/LinMSYYGZL20} applied deliberation networks \cite{DBLP:conf/nips/XiaTWLQYL17} or capsule networks \cite{DBLP:conf/nips/SabourFH17} in the decoder to further optimize the use of visual information. Currently, Multimodal Large Language Models (MLLMs) architectures are widely applied in multimodal tasks \cite{bai2023qwen,yue2024mmmu,li2024llava,huang2024survey,zhu2024benchmarking}; however, their application in MMT remains underexplored. Our approach introduces MLLMs in the field of machine translation for the first time, combined with strong text-to-image models \cite{bolya2023token,DBLP:conf/cvpr/RombachBLEO22} to generate highly relevant, high-quality images from the source text, thereby enhancing translation performance.

\textbf{Image-Free MMT}: Traditional multimodal approaches require annotated images corresponding to input text, which limits their practical applicability. To overcome this limitation, \citealp{DBLP:conf/acl/HitschlerSR16} proposed using target-end image retrieval to aid translation; \citealp{DBLP:conf/ijcnlp/ElliottK17} designed the multi-task learning framework “Imagination,” which breaks down the translation task into learning both translation and visual association representations; \citealp{DBLP:conf/acl/CalixtoRA19} introduced latent variables to estimate the joint distribution of translations and images; \citealp{DBLP:conf/naacl/LongWL21} used Generative Adversarial Networks (GANs) \cite{DBLP:conf/nips/GoodfellowPMXWOCB14} to generate visual representations for translation prediction. Additionally, \citealp{DBLP:conf/acl/0001LZZC23} introduced a visual scene hallucination mechanism to achieve inference-time image-free machine translation. Building on these studies, our approach further enhances translation performance in the absence of image input. The core of our approach includes: 1) eliminating the need for text and image annotation during training, significantly reducing MMT data costs; 2) using consistency training with LSG and VSG to ensure the relevance between source text and generated images, thus improving translation performance; and 3) leveraging the CLIP model to align visual and textual semantic consistency, further reducing noise interference.

\section{Conclusion}
Our IMAGE framework leverages imaginative generation to enhance LLM-based machine translation, providing clearer visual image that improves translation accuracy. By using graph-based supervision to refine scene and relationship clarity, IMAGE outperforms traditional text-only LLM-MT approaches, especially on complex sentences, and pioneers the integration of visual signals to boost translation performance.

\section{Limitation}
Our IMAGE method utilizes imaginative generation to enhance machine translation based on large language models (LLMs), delivering a clearer visual image that significantly boosts translation accuracy. However, the translation capability of our method is primarily limited by the multilingual performance of LLMs. Additionally, our method requires collaborative training of LLMs and Stable Diffusion, which demands greater computational resources.

\bibliography{custom}

\appendix

\section{Data and Training Setting}
\subsection{Dataset Detail}
\label{data_appendix}

\textbf{Multi30K\cite{DBLP:conf/acl/ElliottFSS16}} We evaluate our methods on two standard benchmarks: Multi30K English$\rightarrow$German (En$\rightarrow$De) and English$\rightarrow$French (En$\rightarrow$Fr). Multi30K is a widely used MMT dataset, containing 31,014 images with one English description and the manual translation in German and French. The training and validation sets consist of 29,000 and 1,014 instances, respectively. We reported the results on the Test2016, Test2017, Test2018 and MSCOCO test sets, which includes 1, 000, 1,000, 1071 and 461 instances, respectively.

\textbf{WMT24 test set \cite{kocmi-etal-2024-findings}} To further validate the effectiveness of our framework in general translation,  we also conducted tests on the WMT24 English$\rightarrow$German (En$\rightarrow$De), English$\rightarrow$Chinese (En$\rightarrow$Zh), English$\rightarrow$Czech (En$\rightarrow$Cs), and Eglish$\rightarrow$Hindi (En$\rightarrow$Hi) test sets. Among them, En$\rightarrow$De and En$\rightarrow$Zh are high-resource MT tasks, while En-Cs and En-Hi are low-resource tasks.

\subsection{Training Setting}
\label{off-the-shelf-tool}
Following prior research, we use Mask R-CNN \cite{DBLP:conf/cvpr/TangNHSZ20} as part of a VSG generator\footnote{https://github.com/KaihuaTang/Scene-Graph-Benchmark.pytorch}. For LSG generation, we parse sentences into dependency trees \cite{DBLP:conf/cvpr/00010BT0GZ18} and transform them into scene graphs based on specific rules \cite{DBLP:conf/acl-vl/SchusterKCFM15}. The SIM tool for calculating the similarity between LSG and VSG uses Sentence Transformers\footnote{https://huggingface.co/sentence-transformers/all-MiniLM-L6-v2}~\cite{DBLP:conf/emnlp/ReimersG19}.

The main experiments is conducted on open-source LLMs from the LLaMA2 family\cite{DBLP:journals/corr/abs-2307-09288}. Specifically, we select DreamLLM\cite{DBLP:conf/iclr/DongHPQGYZSZWK024} as our multimodal large language model, which is based on Vicuna-7B \cite{vicuna2023}. The model is trained for 1.5 epochs with a batch size of 16, a peak learning rate of 2e-5 with 3\% warmup ratio.  We use Deepspeed stage 2\cite{DBLP:conf/kdd/RasleyRRH20} to conduct multi-GPU distributed training, with training precision FP16 enabled. For more specific hyperparameters, please refer to our released scripts. For other models used for comparison, such as Llama3-8B and Alpaca , the settings are also the same.

\end{document}